\documentclass[a4paper,twoside]{article}

\usepackage{epsfig}
\usepackage{graphicx}
\usepackage{svg}
\usepackage{array} 
\usepackage{subcaption}
\usepackage{arydshln} 
\usepackage{calc}
\usepackage{amssymb}
\usepackage{amstext}
\usepackage{amsmath}
\usepackage{longtable} 
\usepackage{parcolumns} 
\usepackage{booktabs} 
\usepackage{amsthm}
\usepackage{multicol}
\usepackage{pslatex}
\usepackage{multirow}
\usepackage{apalike}
\usepackage{tabularx}
\usepackage[linesnumbered,vlined]{algorithm2e}
\usepackage[bottom]{footmisc}
\usepackage{listings}
\usepackage{xcolor}
\usepackage{caption}
\usepackage{hyperref}
\usepackage{fancyhdr}
\usepackage{SCITEPRESS}     

\definecolor{mybackground}{rgb}{0.95,0.95,0.95}
\definecolor{mycomment}{rgb}{0.0,0.6,0.0}
\definecolor{mykeyword}{rgb}{0.0,0.0,1.0}
\definecolor{mystring}{rgb}{1.0,0.0,0.0}
\definecolor{myidentifier}{rgb}{0.0,0.0,0.0}
\definecolor{myframe}{rgb}{0.0,0.0,0.0} 

\newcommand{\name}{\textsc{AgentForge}\xspace}

\begin{document}

\title{\name: A Flexible Low-Code Platform for RL Agent Design}

\author{\authorname{Francisco E. Fernandes Jr.\sup{1}\orcidAuthor{0000-0003-2301-8820} and Antti Oulasvirta\sup{1}\orcidAuthor{0000-0002-2498-7837}}
\affiliation{\sup{1}~Department of Information and Communications Engineering, Aalto University, Espoo, Finland}
\email{\{francisco.fernandesjunior, antti.oulasvirta\}@aalto.fi}
}


\abstract{
Developing a reinforcement learning (RL) agent often involves identifying values for numerous parameters, covering the policy, reward function, environment, and agent-internal architecture. Since these parameters are interrelated in complex ways, optimizing them is a black-box problem that proves especially challenging for nonexperts. Although existing optimization-as-a-service platforms (e.g., Vizier and Optuna) can handle such problems, they are impractical for RL systems, since the need for manual user mapping of each parameter to distinct components makes the effort cumbersome. It also requires understanding of the optimization process, limiting the systems' application beyond the machine learning field and restricting access in areas such as cognitive science, which models human decision-making. To tackle these challenges, the paper presents \name, a flexible low-code platform to optimize any parameter set across an RL system. Available at \url{https://github.com/feferna/AgentForge}, it allows an optimization problem to be defined in a few lines of code and handed to any of the interfaced optimizers. With \name, the user can optimize the parameters either \emph{individually} or \emph{jointly}. The paper presents an evaluation of its performance for a challenging vision-based RL problem.
}

\onecolumn \maketitle \normalsize \setcounter{footnote}{0} \vfill

\renewcommand\thefootnote{} 
\footnotetext{This paper has been accepted at the 17th International Conference on Agents and Artificial Intelligence (ICAART 2025).}
\renewcommand\thefootnote{\arabic{footnote}}

\section{\uppercase{Introduction}}
\label{sec:introduction}

Developing reinforcement learning (RL) agents is important not only for advancements in machine learning (ML) but also for fields such as the cognitive sciences, where RL increasingly serves as a computational framework to model human decision-making mechanisms~\cite{eppe_intelligent_2022}. For example, RL can help researchers understand how humans behave under cognitive constraints~\cite{chandramouli2024workflow}. One of the challenging aspects of this development is optimizing a broad array of parameters that influence agent behavior and performance: this process represents a black-box optimization problem. Embodied RL agents, modeled as partially observable Markov decision processes (POMDPs)~\cite{eppe_intelligent_2022}, highlight this complexity in tasks such as robotic navigation~\cite{shahria_comprehensive_2022} and human--computer interaction~\cite{jiang_eyeformer_2024}. In contrast against ML domains wherein the tuning focuses on a smaller set of training-related parameters, RL requires optimizing explainable ones too (e.g., reward weights), to assure of trustworthiness in artificial intelligence (AI) systems.

We set out to give domain experts in cognitive sciences, alongside other fields, the ability to optimize all RL system parameters, either \emph{jointly} or \emph{individually}, as conditions dictate. This is essential, since even small changes in an RL algorithm's implementation, such as its reward clipping, can significantly affect performance~\cite{engstrom_implementation_2020}. In some cases, careful parameter selection can improve performance more than the choice of RL algorithm itself~\cite{andrychowicz_what_2020}. Embodied RL agents in particular employ deep neural networks with numerous parameters highly sensitive to optimization~\cite{fetterman_tune_2023}. These parameters can be roughly categorized into three classes:
\begin{enumerate}
\item \textbf{Agent design} controls internal functions such as the discount factor, entropy coefficient, and observation window size.
\item \textbf{Environment settings} define the task, including the field of view and the reward structure.
\item \textbf{Policy parameters} encompass neural-network architecture, learning rate, and activation functions.
\end{enumerate}

Against this backdrop, we present the ongoing work on \name, a flexible low-code platform designed for experts across disciplines to develop RL systems without having to possess expertise in optimization or ML. This domain-agnostic ability aligns with the needs of cognitive scientists, who often require tailored experiment setups for testing hypotheses about human cognition~\cite{eppe_intelligent_2022,nobandegani_cognitive_2022}. The development of low-code platforms is relevant to many such fields. Also, \name provides for rapid iteration by letting users prototype RL designs, optimize them, and analyze the results. It supports a wide range of RL systems, from simple models to complex embodied agents. Users need only define a custom environment, an objective function, and parameters in two files, where the objective function, which guides the optimization process, can employ any criterion (e.g., average reward per episode). The platform then converts these inputs into an optimization problem, automating training and evaluation. \name's current version includes random search, Bayesian optimization (BayesOpt), and particle swarm optimization (PSO), with flexibility for adding more techniques.

Obviously, parameter optimization is a perennial challenge in ML. Frameworks such as BoTorch~\cite{balandat_botorch_2020}, Optuna~\cite{akiba_optuna_2019}, and Google Vizier~\cite{song_open_2023} represent attempts to address it. However, these tools are targeted at users with advanced ML knowledge and are tricky to apply to RL agents on account of the complexity of defining objective functions for parameters across system components. This shortcoming highlights the need for \emph{low-code solutions}, such as ours, that allow users to design effective agents without having mastered the optimization techniques crucial for handling high-dimensional inputs and dynamic RL environments.

We tested \name by jointly optimizing the parameters of a pixel-based Lunar Lander agent from the Gymnasium library~\cite{towers_gymnasium_2024}, a POMDP in which the agent infers the state from raw pixel values. Since this agent includes parameters from all three categories -- agent, environment, and policy -- it permitted testing \name's ability to handle complex parameter sets. Also, the paper demonstrates how easily an optimization problem can be defined, thus highlighting our platform's simplicity and flexibility.

\section{\uppercase{Related Works}}
\label{sec:related-works}

\begin{table*}[t]
  \caption{Example RL agents and corresponding parameterized modules -- highlighting the diversity of parameter types, this non-exhaustive list emphasizes the complexity of parameter optimization in RL systems}
  \label{tab:agent_types}
  \centering
  \resizebox{\textwidth}{!}{%
  \begin{tabular}{m{6cm}m{5cm}m{5cm}m{4.5cm}}
    \toprule
    \textbf{Agent}/\textbf{env.} & \textbf{Modules} & \textbf{Use case} & \textbf{Typical user} \\
    \midrule
    \raggedright
    Gymnasium toy environments~\cite{towers_gymnasium_2024} & \begin{tabular}{@{}m{5cm}} \raggedright
      - Environment dynamics \\
      - Reward structure \\
      - Policy network \\
      - Value network \\
    \end{tabular} & \begin{tabular}{@{}m{5cm}} \raggedright Achieving the highest reward in a well-defined environment \end{tabular} & \begin{tabular}{@{}m{5cm}} \raggedright - Machine learning researcher \\ - Robotics researcher \end{tabular} \\
    \hline
    \raggedright
    ActiveVision-RL~\cite{shang_active_2023} & \begin{tabular}{@{}m{5cm}}
    \raggedright
      - Observation window size \\
      - Persistence of vision \\
      - Sensorimotor reward \\
      - Q-Network \\
    \end{tabular} & \begin{tabular}{@{}m{5cm}} \raggedright Modeling visual attention in complex environments \end{tabular} & \begin{tabular}{@{}m{5cm}} \raggedright - Machine learning researcher \\ - Robotics researcher \end{tabular}\\
    \hline
    \raggedright
    CRTypist~\cite{shi_crtypist_2024} & \begin{tabular}{@{}m{5cm}}
    \raggedright
      - Foveal vision \\
      - Peripheral vision \\
      - Finger movement \\
      - Working memory \\
      - Reward structure \\
      - Neural networks \\
    \end{tabular} & \begin{tabular}{@{}m{2.5cm}} Simulating touchscreen typing behavior \end{tabular} & \begin{tabular}{@{}m{5cm}} \raggedright - Designer \\ - Accessibility researcher \\ - Cognitive scientist \\ - Behavioral scientist\end{tabular}\\
    \hline
    \raggedright
    Autonomous aircraft landing~\cite{ladosz_autonomous_2024} & \begin{tabular}{@{}m{5cm}}
    \raggedright
      - Descent velocity \\
      - Frame stack \\
      - Image resolution \\
      - Q-Network \\
      - Reward structure \\
    \end{tabular} & \begin{tabular}{@{}m{5cm}} \raggedright Automating aircraft landings with vision-based reinforcement learning \end{tabular} & \begin{tabular}{@{}m{5cm}} \raggedright - Machine learning researcher \\ - Robotics researcher \end{tabular} \\
    \hline
    \raggedright
    EyeFormer~\cite{jiang_eyeformer_2024} & \begin{tabular}{@{}m{5cm}}
    \raggedright
      - Inhibition of return (IoR) \\
      - Vision encoder \\
      - Fixation decoder \\
      - Optimization constraints \\
      - Neural networks \\
      - Reward structure \\
    \end{tabular} & \begin{tabular}{@{}m{5cm}} \raggedright Modeling visual attention and eye-movement behavior for UI optimization  \end{tabular} & \begin{tabular}{@{}m{5cm}}
    \raggedright
      - Designer \\
      - Accessibility researcher \\
      - Cognitive scientist\\
      - Behavioral scientist
    \end{tabular} \\
    \hline
  \end{tabular}
  }
\end{table*}

Among the many fields showing extensive use of RL are robotics, cognitive science, and human--computer interaction in general. Robotics focuses on optimizing real-world sensorimotor interactions~\cite{shang_active_2023}, while cognitive scientists model human decision-making~\cite{chandramouli2024workflow}. Table~\ref{tab:agent_types} highlights the breadth of tasks and needs in RL development. Across the board, these share the challenge of optimizing large parameter spaces and adapting to dynamic environments, a goal that drives demand for flexible, low-code platforms that simplify RL development across disciplines.

The work in response forms part of the field known as AutoRL, aiming to automate RL algorithm and parameter optimization. While already showing clear progress, AutoRL is in its infancy~\cite{parker-holder_automated_2022,afshar_automated_2022}. Platforms such as AutoRL-Sim~\cite{souza_autorl-sim_2024} and methods that use genetic algorithms~\cite{cardenoso_fernandez_parameters_2018} support simple RL tasks but lack support for more complex real-world problems with diverse parameters across agents, environments, and policies. Among the optimization techniques commonly employed in AutoRL are random search, BayesOpt, and PSO, with the choice of technique depending on task-specific requirements. Furthermore, while such state-of-the-art RL solvers as Stable-Baselines3~\cite{raffin_stable-baselines3_2021} and CleanRL~\cite{huang_cleanrl_2022} offer powerful tools for agent training, they demand deep knowledge of RL's internals, plus extensive coding, and are not suitable for parameter optimization.

Platforms of \name's ilk can address these challenges by simplifying RL agent design and providing automated optimization for diverse needs. Its provision for random search, BayesOpt, and PSO furnishes a low-code interface for joint parameter optimization, whereby users become able to compare and apply strategies for various RL agents with ease.

\section{\uppercase{The Proposed Framework}}
\label{sec:proposed-framework}

\subsection{General Characterization}

For enabling flexible low-code specification of an optimization problem in RL, the following engineering objective guided the design of \name:
\begin{itemize}
    \item \textbf{A low-code setting:} Users provide simple config files that state optimization parameters and performance-evaluation methods. This approach lets users exploit the platform without deep technical expertise.
    \item \textbf{Integration:} Support for custom RL environments compatible with the Gymnasium library enables use across diverse application scenarios.
    \item \textbf{Flexibility:} Thanks to its multiple optimization methods, users can easily swap and set up methods through config files without changing the core code. This simplifies adaptation to various problem requirements.
    \item \textbf{Parallelization:} Allowing concurrent evaluation of multiple solutions reduces optimization time. By using Optuna's parallelization and our custom methods, the platform can explore many configurations simultaneously, which is especially beneficial for large-scale RL problems.
\end{itemize}

\begin{figure*}[!t]
    \centering
    \includegraphics[width=\textwidth]{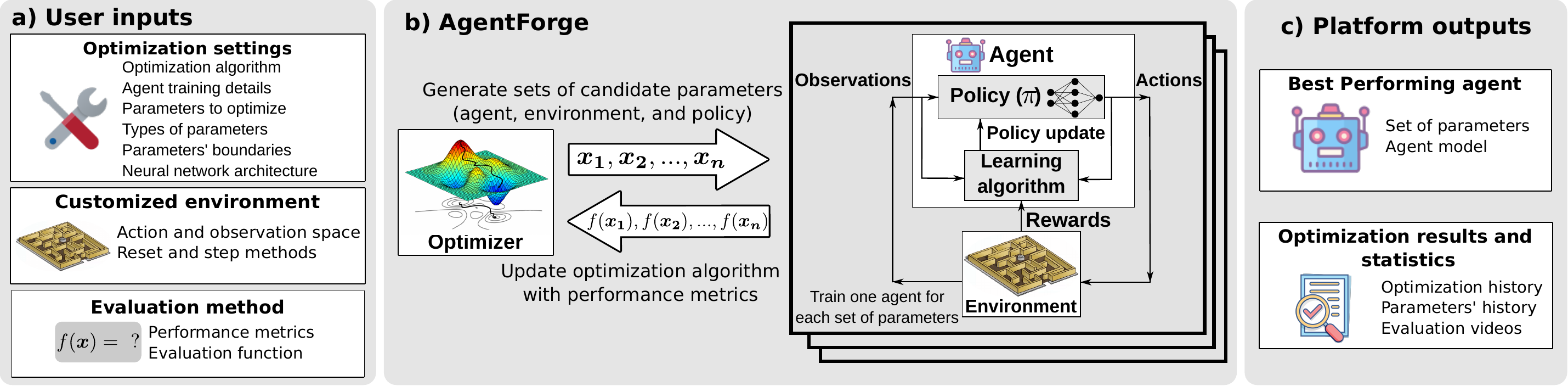}
    \caption{The \textit{\name} platform in overview. a) The inputs consist of three elements: optimization settings (including search parameters), a customized environment, and an evaluation method. b) \textit{\name} executes joint optimization of RL parameters through a two-loop process. c) Then, \textit{\name} provides the user with the best-performing agent along with the optimization results and performance statistics.}
    \label{fig:optimization-overview}
\end{figure*}

The overall optimization process comprises two loops, illustrated in Figure~\ref{fig:optimization-overview}. The outer loop optimizes parameters via the chosen method, while the inner one trains the RL agent through environment interactions and performance evaluations. With the next few sections, we give more details on using the framework and on its optimization algorithms.

\vspace{10pt}
\begin{figure}[b]
\begin{lstlisting}[language=Python]
...
# Training parameters
gae_lambda:
    type: default
    searchable: true
    integer: false
    user_preference: 0.9
    start: 0.9
    stop: 0.95
...
def user_train_environment(seed, n_train_envs, **env_kwargs):
    """
    Set up RL environment for training.
    Args:
    - seed (int): Random seed.
    - n_train_envs (int): Number of training envs.
    - **env_kwargs: Additional args for env creation.
    Returns:
    - train_env (VecEnv): Configured training environment.
    """ 
def user_evaluate_policy(seed, model, **env_kwargs):
    """
    Evaluate the policy with a separate evaluation environment.
    Args:
    - model (BaseAlgorithm): Model to eval.
    Returns:
    - mean_objective_value (float): Mean objective value obtained during evaluation.
    - extra_info (dict): Additional metrics defined by the user.
    """
\end{lstlisting}
\captionof{lstlisting}{Sample code snippet illustrating parameter definition, creation of the environment, and evaluation setup.}\label{code:evaluation-function}
\end{figure}
\vspace{10pt}
\setlength{\parindent}{15pt}

\subsection{Specifying the Optimization Task}
\label{subsec:user-configuration}
To use \name effectively, the user must furnish three key components: configuration settings, a customized environment, and an evaluation method.

Users start by providing a YAML configuration file to define the optimization setup. This file specifies the parameters to be optimized, each categorized as agent, environment, or policy, along with the type for each (integer or float) and value ranges. It also includes settings for agent training hyperparameters, such as learning rates and discount factors, and the choice of optimization algorithm: in the current implementation random search, BayesOpt, or PSO. Additionally, \name supports optimizing neural network architectures for policy and value networks. thereby letting users set the range for the number of layers and of neurons per layer. Thus it enables a thorough search for optimal structures.

Next, via a Python script, the user implements a custom RL environment compatible with the Stable-Baselines3 library. It must adhere to the standard interface, defining action and observation spaces, and include methods for resetting and stepping through the environment. For this, the user writes a function that can create a training environment for the agent. While \name currently supports only the popular Python library Stable-Baselines3, future versions may extend compatibility to other libraries.

Finally, users must implement an evaluation method, again in a Python script, to assess the trained agent's performance. The function should calculate and report metrics for factors such as reward, success rate, etc., to make sure that \name can collect and use these values during optimization. Optionally, they may also provide a function that records a video of the agent's performance over a specified number of episodes in the environment.

These three components are summarized in Figure~\ref{fig:optimization-overview} and in Listing~\ref{code:evaluation-function}.

\subsection{Optimization Algorithms}
\name supports several popular optimization algorithms. While random search, BayesOpt, and PSO form the set of default options, the framework is flexible; it can be extended to include other methods, for various RL scenarios.

\begin{itemize}
    \item \textbf{Random search} is a simple baseline method that randomly samples the parameter space. Useful for initial comparisons in testing of more sophisticated algorithms, it guarantees rich exploration but is less efficient at finding optimal regions.
    \item \textbf{BayesOpt} is a structured approach using probabilistic models (e.g., Gaussian processes) to focus on promising areas of the parameter space. Our platform uses a tree-structured Parzen estimator (TPE), well suited to high-dimensional spaces~\cite{bergstra_algorithms_2011}. It efficiently balances exploration and exploitation for fast convergence.
    \item \textbf{PSO} is a population-based approach suitable for continuous optimization. Its stochastic nature aids in escaping local optima. We use inertia ($w$), cognitive ($c_1$), and social ($c_2$) weights of $0.9694$, $0.099381$, and $0.099381$, respectively, for better exploration of high-dimensional spaces, as recommended by~\cite{oldewage_movement_2020}.
\end{itemize}

In \name, Random search and BayesOpt are implemented with Optuna, while PSO is custom-designed. The mechanism of letting users select and configure optimization strategies through a user-defined configuration file (see the previous subsection) renders the framework adaptable to various applications. 

\subsection{Parallel Evaluation of Trials}
Taking advantage of parallel evaluation improves the efficiency and speeds up optimization. By running each trial as an independent Python process, we avoid issues with Python's Global Interpreter Lock. For all optimization methods, parameter sets are distributed across nodes for evaluation, with multiple processes accessing a centralized database to get a new set for each trial. Parallelization is crucial since each trial involves training agents for millions of timesteps, which often takes hours. By utilizing parallel computing, the platform significantly reduces the time required for a full optimization loop, thus facilitating extensive RL experimentation.

\section{\uppercase{Evaluation}}
\label{sec:evaluation}

\begin{table}[t]
  \caption{General optimization configuration, applied across all experiments}
  \label{tab:general_optimization_configuration}
  \centering
  \resizebox{\columnwidth}{!}{%
  \begin{tabular}{ll}
    \toprule
    \multicolumn{2}{c}{\textbf{Training setup for each trial}} \\
    \midrule
    Number of training timesteps & 5,000,000 \\
    Evaluation frequency & Every 10 batches \\
    Maximum steps per episode & 1024 \\
    Vectorized environments for training & 10 \\
    Training algorithm & PPO \\
    Policy architecture & CNN \\
    \midrule
    \multicolumn{2}{c}{\textbf{Random search}} \\
    \midrule
    Number of trials & 400 \\
    Pruning of trials & After 50 trials \\
    Pruning strategy & MedianPruner \\
    Wait before trial is pruned & At least 48 evaluations \\
    \midrule
    \multicolumn{2}{c}{\textbf{BayesOpt}} \\
    \midrule
    Number of trials & 400 \\
    Number of initial random trials & 50 \\
    Number of improvement candidates expected & 80 \\
    Multivariate TPE & Enabled \\
    Pruning of trials & After 50 trials \\
    Pruning strategy & MedianPruner \\
    Wait before trial is pruned & At least 48 evaluations \\
    \midrule
    \multicolumn{2}{c}{\textbf{PSO}} \\
    \midrule
    Number of generations & 20 \\
    Population size & 20 \\
    Inertia weight ($w$) & 0.9694 \\
    Cognitive component weight ($c_1$) & 0.099381 \\
    Social component weight ($c_2$) & 0.099381 \\
    \bottomrule
  \end{tabular}
  }
\end{table}

We evaluated \name on the basis of \emph{agent performance}, judged in terms of the maximum mean reward achieved by a pixel-based POMDP agent.

\subsection{A Pixel-Based POMDP Agent}
Our evaluation of \name used a pixel-based version of the Lunar Lander environment from the Gymnasium library~\cite{towers_gymnasium_2024}, a POMDP wherein the agent relies on raw-pixel observations instead of state-based inputs. The agent processes four stacked frames as visual input and uses continuous actions to control the main engine and lateral boosters. Reward structure remains unchanged from the original environment's, with the task considered solved when the agent reaches a reward of at least 200 in a single episode.

This experiment demonstrated \name's flexibility in optimizing parameters across all three categories (agent, environment, and policy), including neural network architectures. The breadth of optimization attests to \name's capacity to handle complex RL configurations effectively. Below, we provide details of its parameter optimization.

\subsection{Optimization Configuration}
The optimization configuration for all experiments, covering the parameters for individual training trials and the random search, BayesOpt, and PSO methods, is detailed in Table \ref{tab:general_optimization_configuration}. Each trial involved training for five million timesteps, with performance evaluations after every 10 batches. This setup used 10 parallel, vectorized environments with the proximal policy optimization (PPO) algorithm and a convolutional neural network (CNN) feature extractor. Agent performance was measured as the mean reward over 300 independent episodes. The table includes all the optimization settings: the number of trials, pruning strategies, and algorithm-specific parameters. All configurations can be easily customized through the user-defined configuration file introduced in Section~\ref{sec:proposed-framework}.

\subsection{The Parameters Optimized}
We optimized the parameters listed in Table~\ref{tab:parameters_optimized-results}. The environment parameter chosen for optimization here was the size of the field of view (in pixels). This parameter determines how much the original environment window is downsampled into a square window suitable for the feature extractor. To define its dimensions, only one value is needed. Further parameters optimized were the discount factor ($\gamma$), generalized advantage estimation ($\lambda$), learning rate, number of epochs, entropy coefficient, and clipping range. These are critical for training because they influence reward discounting, learning stability, and the efficiency of policy updates. Also tuned were neural network architecture parameters, such as the activation function, number of layers, and number of neurons per layer for the policy and value networks both. The activation function was defined as a float in the range [0.0, 1.0], with Tanh used for values below 0.5 and ReLU for those of 0.5 or above.

\begin{table*}[t!]
  \caption{The parameters optimized and the best values found for the pixel-based Lunar Lander agent -- the wide range of parameters being jointly optimized highlights the flexibility of \name}
  \label{tab:parameters_optimized-results}
  \centering
  \resizebox{\textwidth}{!}{%
  \begin{tabular}{lcccccc}
    \toprule
    \textbf{Parameter} & \textbf{Type} & \textbf{Category} & \textbf{Range} & \textbf{Random search} & \textbf{BayesOpt} & \textbf{PSO} \\
    \midrule
    Field of view's size (pixels) & Integer & Environment & [40, 92] & 71 & 92 & 92 \\
    Discount factor ($\gamma$) & Float & Agent & [0.4, 0.8] & 0.7934 & 0.7984 & 0.8 \\
    Generalized advantage estimation ($\lambda$) & Float & Agent & [0.9, 0.95] & 0.9433 & 0.9299 & 0.95 \\
    Learning rate & Float & Agent & [$3.5\cdot 10^{-4}$, $3.5\cdot 10^{-3}$] & 0.0006 & 0.0034 & 0.0035 \\
    Number of epochs & Integer & Agent & [3, 10] & 6 & 3 & 5 \\ 
    Entropy coefficient & Float & Policy & [0.01, 0.1] & 0.0284 & 0.0279 & 0.1 \\
    Clipping range & Float & Policy & [0.01, 0.3] & 0.1407 & 0.0178 & 0.3 \\
    Activation function & Float & Policy & [0.0, 1.0] & 0.5478 & 0.9259 & 1.0 \\
    No. of layers in policy & Integer & Policy & [1, 4] & 1 & 3 & 4 \\ 
    No. of neurons per layer in policy & Integer & Policy & [64, 128] & 83 & 89 & 127 \\
    No. of layers in value network & Integer & Policy & [2, 4] & 4 & 2 & 4 \\
    No. of neurons per layer in value & Integer & Policy & [64, 128] & 100 & 89 & 128 \\
    \midrule
    \textbf{Mean reward} & & & & 84.26 & \textbf{172.43} & 119.19 \\
    \bottomrule
  \end{tabular}
  }
\end{table*}

\subsection{Joint vs. Individual-Parameter Optimization}
The BayesOpt results presented below demonstrate the flexibility of \name well. We report these for three parameter-optimization configurations:

\begin{itemize}
    \item \textbf{Joint optimization with NAS:} Simultaneous optimization of all parameters listed in Table~\ref{tab:parameters_optimized-results}, addressing the policy and value networks' architecture
    \item \textbf{Optimization without NAS:} Optimizing all parameters except for the network architecture (activation function, number of layers, and neuron count), which remain at their stable-baselines3 defaults
    \item \textbf{NAS only:} Optimization restricted to the network architecture while all other parameters are fixed at their Stable-Baselines3 defaults
\end{itemize}

\section{\uppercase{Results}}
\label{sec:results}

\begin{figure}
    \centering
    \includegraphics[width=\columnwidth]{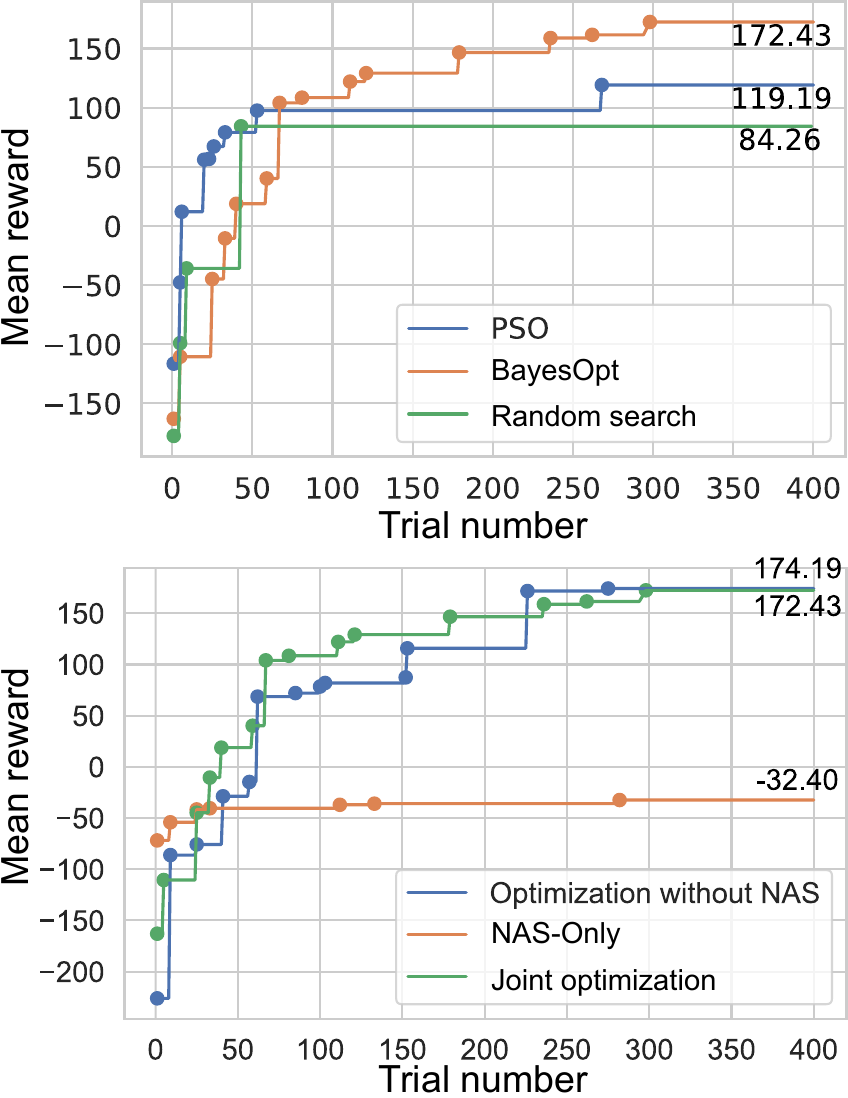}
    \caption{Mean reward per episode for the pixel-based Lunar Lander agent under different optimization strategies. \textit{Top:} Joint optimization of parameters and NAS using random search, BayesOpt, and PSO. \textit{Bottom:} A plot of joint vs. individual-parameter optimization using BayesOpt, with and without NAS, and NAS-only optimization.}
    \label{fig:results-lunar-lander}
\end{figure}

\subsection{Results from Joint Optimization}
The upper pane of Figure~\ref{fig:results-lunar-lander} shows the evolution of the mean reward in joint optimization over these episodes for each trial under random search, BayesOpt, and PSO. One can see that BayesOpt achieved the highest cumulative rewards while PSO showed the fastest convergence. Table~\ref{tab:parameters_optimized-results} lists the best parameters found for this agent.

BayesOpt consistently outperformed random search and PSO in terms of mean cumulative reward, achieving a mean reward of 172.43. Though PSO reached a lower final reward (mean reward: 119.19), it demonstrated the fastest convergence. Random search lagged behind with a mean reward of 84.26, serving as a baseline for comparison. Notably, PSO reached approximately 69\% of BayesOpt's mean reward, while random search represented about 49\% of BayesOpt's level. These results spotlight the effectiveness of BayesOpt but also demonstrate \name's flexibility in integrating multiple optimization algorithms, which supports users' selection of the best approach for their specific RL agents.

\subsection{Results of the Joint vs. Parameter-Wise Optimization}
Figure~\ref{fig:results-lunar-lander}'s lower pane presents the results of \emph{joint} versus \emph{individual-parameter} optimization. Joint optimization with and without NAS achieved mean episode rewards of $172.43$ and $174.19$, respectively, while the NAS-alone condition yielded a mean episode reward of $-32.40$. These findings suggest that network architecture is not the most critical parameter set to optimize for our agent. However, users short on time can obtain satisfactory results by opting for complete joint optimization.

\subsection{Platform Overhead}
\name introduces negligible overhead relative to the time required for evaluating RL agents. A single candidate's evaluation takes roughly three hours on a consumer-grade Nvidia RTX 4090 GPU, making any additional delays from the optimization process insignificant. However, parallel evaluation is vital; running 400 trials serially would take about 50 days. By evaluating eight trials in parallel, we can complete the same 400 trials in just seven days.

\section{\uppercase{Discussion}}
\label{sec:discussion}
Our results affirm \name's effectiveness as a flexible, low-code platform for RL agent design and optimization. It addresses the challenge of optimizing interrelated parameters in large numbers across the agent, environment, and policy classes by treating this as a black-box optimization problem. We found evidence that parameter optimization on its own can improve the RL agent's performance, without reliance on changes to the training algorithms. All three optimization methods currently available with \name enhanced performance in our test case, with BayesOpt arriving at the highest episode mean reward observed. However, the platform's key strength lies in its support for joint and individual-level optimization of RL agents' parameters and neural network architectures. This crystallizes the platform's flexibility for strongly supporting the study of critical interplay among various aspects of RL agent design.

In principle, \name's integration of various optimization methods allows users to compare approaches and can streamline the prototyping and fine-tuning of RL agents. By configuring optimization settings through a simple YAML file and a Python script, they can quickly customize their agents without extensive coding. Thus helping pare back the typical trial-and-error process in RL experimentation, this frees more of practitioners' time for refining their problem models. While our results are encouraging, further testing, with user studies and different RL systems, will be necessary to validate \name's broader impact. Still, it already offers a flexible solution to effectively tackle the unique challenges of RL systems, overcoming several issues that practitioners might face when using other platforms.

\section{\uppercase{Conclusions and Future Work}}
\label{sec:conclusion}

With this short paper, we provided a glimpse at \textit{\name}, which simplifies joint optimization of RL parameters, accelerating agent development -- especially for users with little optimization experience -- through flexible optimization and easier design. Evaluation with a pixel-based POMDP agent settings proved illuminating. The setup was simple from a low-code perspective yet proved able to support finding a suitable balance (e.g., BayesOpt delivered the best optimization results, while PSO showed faster convergence but requires further tuning). The platform also gives users control over their experiment setups: it demonstrated good flexibility for both joint and individual parameters' optimization.

This version does display several limitations, which will be addressed in a follow-up paper. For instance, we are developing a graphical interface to enable a comprehensive user study evaluating the ease of use of our solution in comparison to preexisting ones such as AutoRL-Sim. Also, the current interface is designed to assist cognitive and behavioral scientists in using RL to model human decision-making processes more effectively. We plan to test use cases additional to these, probe more agent designs, and fully address the challenge of managing RL training uncertainty during optimization.

\section*{\uppercase{Acknowledgements}}
This work was supported by the ERC AdG project Artificial User (101141916) and the Research Council of Finland (under the flagship program of the Finnish Center for Artificial Intelligence, FCAI). The calculations were performed via computer resources provided by the Aalto University School of Science project Science-IT. The authors also acknowledge Finland's CSC – IT Center for Science Ltd. for providing generous computational resources.

\bibliographystyle{apalike}
{\small
\bibliography{references}}

\end{document}